\documentclass[b5paper,10pt,abstractoff,DIV=calc,headings=normal]{scrartcl}
\usepackage[english]{babel}
\usepackage{lmodern}
\usepackage{microtype}
\usepackage{amsmath,amssymb,amsthm,amsfonts,graphicx,etoolbox,scrlayer-scrpage}
\usepackage[noblocks]{authblk}
\makeatletter
\patchcmd{\@maketitle}{\huge}{\Large}{}{}
\patchcmd{\abstract}{\quotation}{}{}{}
\AtBeginEnvironment{abstract}{\noindent\ignorespaces}
\AtEndEnvironment{abstract}{\par\mbox{}}
\newcommand{\shortauthor}{}
\newcommand{\shorttitle}{\@title}
\makeatother
\setkomafont{author}{\small}
\setkomafont{title}{\rmfamily\bfseries}
\setkomafont{disposition}{\rmfamily\bfseries}

\def\AMS#1{\par\noindent \textbf{AMS subject classification: }#1\par}
\newcommand{\acknowledgements}{\par\mbox{}\par\noindent\textbf{Acknowledgements: }}
\newcommand{\keywords}[1]{\par\noindent\textbf{Keywords: }#1}
\rehead[]{\shortauthor}\lohead[]{\shorttitle}\lehead[]{324}\rohead[]{\shortauthor}
\theoremstyle{plain}

\theoremstyle{definition}

\theoremstyle{remark}

\renewenvironment{abstract}{\bigskip\noindent\begin{minipage}{\textwidth}\setlength{\parindent}{15pt}\paragraph{Abstract:}}{\end{minipage}}

\usepackage{amssymb}   
\usepackage{amsthm}    
\usepackage{amsmath}
\usepackage{thmtools}  
\usepackage{mathtools} 
\usepackage{mathrsfs}  
\usepackage{enumitem}  

\usepackage{graphicx}  
\usepackage{adjustbox}
\graphicspath{{figurer/}}
\usepackage{booktabs, multirow}  
\usepackage{babel}     
\usepackage{csquotes}  
\usepackage{textcomp}  
\usepackage{listings}  
\usepackage{caption}
\usepackage{subcaption}

\usepackage{accents}
\newcommand{\thickbar}{} 
\DeclareRobustCommand*\thickbar[1]{\accentset{\rule{.35em}{.65pt}}{#1}}

\newcommand{\abs}[1]{\left| #1 \right|}%
\usepackage{hyperref}
\usepackage{varioref}
\usepackage{cleveref}

\usepackage{xcolor,colortbl}  
\definecolor{col_ind}{HTML}{E81A20}
\definecolor{col_emp}{HTML}{FF7F00}
\definecolor{col_par}{HTML}{FFDF0F}
\definecolor{col_gen}{HTML}{33A02C}
\definecolor{col_sep}{HTML}{A6CEE3}
\definecolor{col_sur}{HTML}{1F78B4}

\newcommand{\RR}{\mathbb{R}}   
\newcommand{\E}{\mathbb{E}}   

\newcommand{\x}{\boldsymbol{x}}

\newcommand{\0}{\boldsymbol{0}}

\newcommand{\xsb}{{\boldsymbol{x}_{\thickbar{\mathcal{S}}}}}

\newcommand{\xs}{{\boldsymbol{x}_{\mathcal{S}}}}
\newcommand{\xss}{{\boldsymbol{x}_{\mathcal{S}}^*}}

\newcommand{\s}{{\mathcal{S}}}
\newcommand{\M}{{\mathcal{M}}}
\newcommand{\pow}{{\mathcal{P}}}

\newcommand{\bphi}{{\boldsymbol{\phi}}}

\newcommand{\parametric}{\texttt{parametric}}

\newcommand{\Gaussian}{\texttt{Gaussian}}

\DeclareMathOperator*{\argmin}{arg\,min}

\let\OLDthebibliography\thebibliography
\renewcommand\thebibliography[1]{
	\OLDthebibliography{#1}
	\setlength{\parskip}{0pt}
	\setlength{\itemsep}{0pt plus 0.3ex}
}

\begin{document}


\renewcommand{\shortauthor}{Lars H. B. Olsen}
\renewcommand{\shorttitle}{Shapley Value Explanations}

\title{{\fontsize{13.4pt}{13.4pt}\selectfont Precision of Individual Shapley Value Explanations}}

\author[1,2]{Lars Henry Berge Olsen\thanks{\url{lholsen@math.uio.no}}}
\affil[1]{Department of Mathematics, University of Oslo, Norway}
\affil[2]{The Alan Turing Institute, London, United Kingdom\vspace{-2ex}}
\vspace{-2ex}
\maketitle
\vspace{-2ex}

\begin{abstract}
Shapley values are extensively used in explainable artificial intelligence (XAI) as a framework to explain predictions made by complex machine learning (ML) models. In this work, we focus on conditional Shapley values for predictive models fitted to tabular data and explain the prediction $f(\x^{*})$ for a single observation $\x^{*}$ at the time. Numerous Shapley value estimation methods have been proposed and empirically compared on an average basis in the XAI literature. However, less focus has been devoted to analyzing the precision of the Shapley value explanations on an individual basis. We extend our work in \cite{Olsen2023} by demonstrating and discussing that the explanations are systematically less precise for observations on the outer region of the training data distribution for all used estimation methods. This is expected from a statistical point of view, but to the best of our knowledge, it has not been systematically addressed in the Shapley value literature. This is crucial knowledge for Shapley values practitioners, who should be more careful in applying these observations' corresponding Shapley value explanations.
\end{abstract} 


\smallskip
\AMS{62D10, 62E17, 62G05, 62G07, 68T01, 91A12.} 


\section{Introduction}
Complex ML models often obtain accurate predictions for supervised learning problems in numerous fields but at the cost of interpretability. Not understanding the input's influence on the ML model's output is a significant drawback; hence, the XAI field has proposed several types of \textit{post hoc} explanation frameworks \cite{Molnar}. One of the most commonly used explanation frameworks is \textit{Shapely values}, a promising explanation methodology with desirable and unique theoretical properties and a solid mathematical foundation \cite{aas2021explaining, lundberg2017unified, shapley1953value}. 

Shapley values originated in cooperative game theory as a solution concept of how to fairly divide a payout of a game between the players based on their contribution, but it was reintroduced as an explanation framework in XAI by \cite{lundberg2017unified, strumbelj2010efficient}. 
It is most commonly used as a \textit{model-agnostic} explanation framework with  \textit{local explanations}. Model-agnostic means that Shapley values do not rely on model internals and can explain predictions made by any predictive model $f$. Local explanation means that Shapley values explain the local model behavior for a single prediction $f(\x^*)$ by providing feature importance scores and not the global model behavior across all data instances. 

We focus on \textit{conditional} Shapley values, which take feature dependencies into consideration, in contrast to \textit{marginal} Shapley values. A disadvantage of the conditional Shapley values, compared to the marginal counterpart, is that they require the estimation/modeling of non-trivial conditional expectations. There is an ongoing debate about when to use the two versions \cite{chen2022algorithms}, and \cite{Olsen2023} provides an overview of possible estimation methods. Throughout this article, we refer to conditional Shapley values when we discuss Shapley values.

We focus on the supervised learning setting where we aim to explain predictions made by a model $f(\boldsymbol{x})$ trained on $\mathcal{X} = \{\boldsymbol{x}^{[i]}, y^{[i]}\}_{i = 1}^{N_\text{train}}$, where $\boldsymbol{x}^{[i]}$ is an $M$-dimensional feature vector, $y^{[i]}$ is a univariate response, and $N_\text{train}$ is the number of training observations. More specifically, we want to explain individual predictions $f(\boldsymbol{x}^*)$ for specific observations $\x^*$ using Shapley values. We demonstrate and discuss that the explanations will be less precise for test observations in the outer region of the training distribution. Thus, Shapley value practitioners should be more careful when using these explanations.

\vspace{-1.65ex} \section{Shapley Values: Theory and Estimation Methods} \vspace{-1.65ex}
Shapley values are a solution concept of how to fairly divide the payout of a cooperative game $v:\mathcal{P}(\M) \mapsto \RR$ among the $M$ players based on four fairness axioms \cite{shapley1953value}. Here $\M = \{1,2,\dots,M\}$ denotes the set of all players, $\mathcal{P}(\M)$ is the power set, that is, the set of all subsets of $\M$, and $v(\s)$ maps a subset of players $\s \in \pow(\M)$, also called a coalition, to a real number representing their contribution in the game. In XAI, we treat the features $\x^*$ as the players, the predictive model $f$ (indirectly) as the game, and the prediction $f(\x^*)$ as the payout to be fairly distributed onto the features. Furthermore, we call $v(\s)$ the \textit{contribution function}. See \cite{aas2021explaining, lundberg2017unified} for why the four fairness axioms give Shapley values desirable properties in the model explanation setting. 

\begin{figure}[ht]
	\centering
        \vspace{-1ex}
	\centerline{\includegraphics[width=1\textwidth]{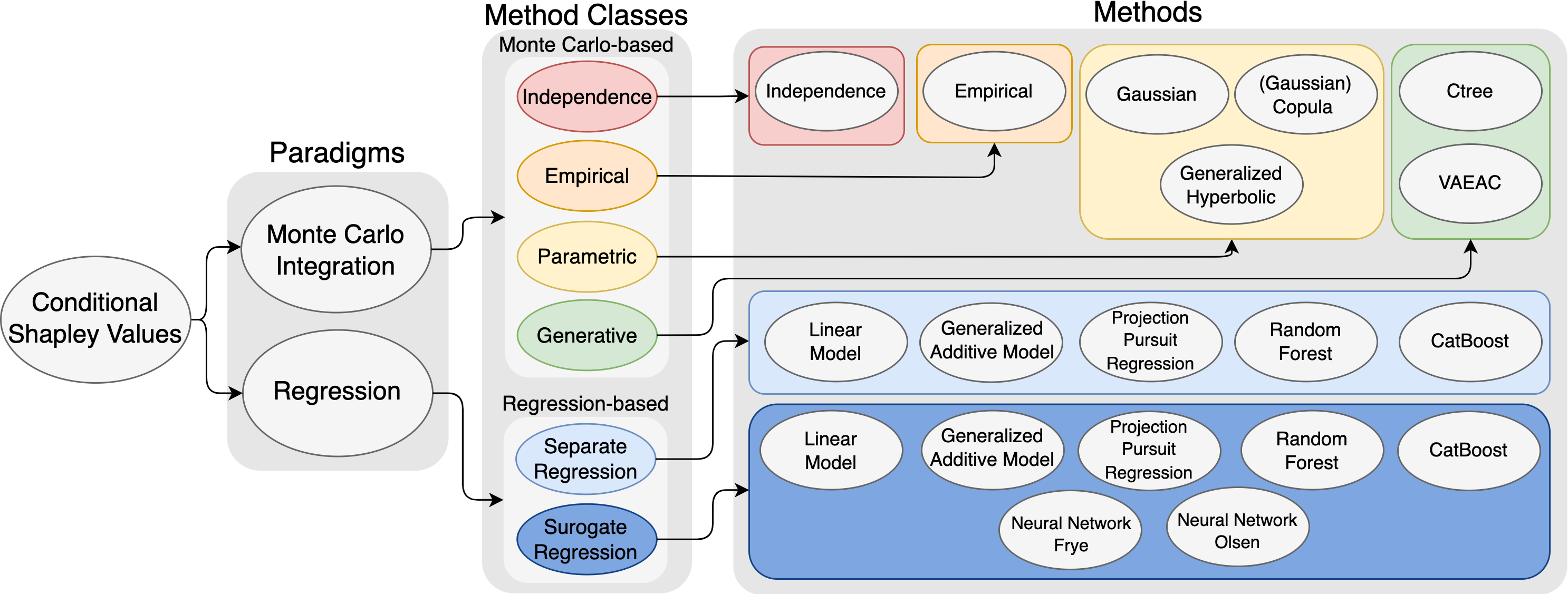}}
	\captionsetup{aboveskip=5pt} 
	\captionsetup{belowskip=-16.5pt} 
	\caption{{\small Schematic overview of the paradigms, method classes, and estimation methods used to compute the conditional Shapley value explanations.}}
	\label{fig:Shapley_Map}
\end{figure}

The Shapley values $\phi_j = \phi_j(v)$ assigned to each feature $j$, for $j = 1, \dots, M$ are given by $\phi_j = \sum_{\mathcal{S} \subseteq \mathcal{M} \backslash \{j\}} \tfrac{|\mathcal{S}|!(M-|\mathcal{S}|-1)!}{M!}\left(v(\mathcal{S} \cup \{j\}) - v(S) \right)$, where $|\mathcal{S}|$ is the number of features in coalition $\s$. Each Shapley value is a weighted average of the feature’s marginal contribution to each coalition $\mathcal{S}$. A common choice for the contribution function in XAI, see \cite{aas2021explaining, covert2021explaining, lundberg2017unified, Olsen2023}, is $v(\mathcal{S}) = v(\mathcal{S}; \x^*) = \E\left[ f(\boldsymbol{x}) | \boldsymbol{x}_{\mathcal{S}} = \boldsymbol{x}_{\mathcal{S}}^* \right] = \E\left[ f(\boldsymbol{x}_{\thickbar{\mathcal{S}}}, \boldsymbol{x}_{\mathcal{S}}) | \boldsymbol{x}_{\mathcal{S}} = \boldsymbol{x}_{\mathcal{S}}^* \right]$, where $\boldsymbol{x}_{\mathcal{S}} = \{x_j:j \in \mathcal{S}\}$ denotes the features in $\mathcal{S}$ and $\boldsymbol{x}_{\thickbar{\mathcal{S}}} = \{x_j:j \in \thickbar{\mathcal{S}}\}$ denotes the features outside the coalition $\mathcal{S}$, that is, $\thickbar{\mathcal{S}} = \mathcal{M}\backslash\mathcal{S}$. One of the desirable Shapley value properties is that $\phi_j^*$ describes the importance of the $j$th feature in the prediction $f(\boldsymbol{x}^*) = \phi_0 + \sum_{j=1}^M\phi_j^*$, where $\phi_0 = \E \left[f(\boldsymbol{x})\right] \approx \bar{y}_\text{train}$. That is, the sum of the Shapley values $\bphi^*$ explains the difference between the prediction $f(\boldsymbol{x}^*)$ and the global average prediction. Note that the Shapley values can be negative.

In \Cref{fig:Shapley_Map}, we provide a schematic overview of the methods used to estimate $v(\s)$ by $\hat{v}(\s)$. We group the methods into six method classes based on their characteristics in accordance with \cite{Olsen2023}. That is, if the methods (implicitly) assume feature independence, use empirical estimates, parametric assumptions, generative methods, or separate/surrogate regression models. We further group the six method classes into two paradigms.

The first one uses Monte Carlo integration, i.e., $\hat{v}(\mathcal{S}) = \frac{1}{K} \sum_{k=1}^K f(\boldsymbol{x}_{\thickbar{\mathcal{S}}}^{(k)}, \boldsymbol{x}_{\mathcal{S}}^*)$, where $K$ is the number of Monte Carlo samples, $f$ is the predictive model, and $\boldsymbol{x}_{\thickbar{\mathcal{S}}}^{(k)} \sim p(\boldsymbol{x}_{\thickbar{\mathcal{S}}} | \boldsymbol{x}_{\mathcal{S}} = \boldsymbol{x}_{\mathcal{S}}^*)$ are the generated Monte Carlo samples, for $k=1,2,\dots,K$. These samples must closely follow the (generally unknown) true conditional distribution of the data to yield accurate Shapley values.

The second paradigm uses that $v(\s)$ is the minimizer of the mean squared error (MSE) loss function, i.e., $v(\s) = \argmin_c \E\left[(f(\xsb, \xs) - c)^2 | \xs = \xss\right]$. Thus, any regression model $g_\s(\xs)$, fitted to the MSE loss function, will approximate $v(\s)$ and yield an alternative estimator $\hat{v}(\s) = g_\s(\xs)$. The accuracy of $\hat{v}(\s)$ depends on, e.g., the form of the predictive model $f(\x)$ and the flexibility of the regression model $g_\s(\xs)$. We can either train a separate regression model $g_\s(\xs)$ for each $\s \in \pow(\M)$ or a single surrogate regression model $g(\tilde{\x}_\s)$ which approximates the contribution function $v(\s)$ for all $\s \in \pow(\M)$ simultaneously. In \cite{Olsen2023}, we thoroughly discuss the notation, method (implementation) details, and how the methods estimate the true conditional distribution/expectation, and we provide extensive recommendations for which method (or methods) to use in different situations.

\vspace{-1.4ex} \section{Simulation Study: Results and Discussion} \vspace{-1.4ex}
We focus on the \texttt{gam\textunderscore more\textunderscore interactions} experiment in \cite{Olsen2023}, but we obtain similar results for the other experiments in \cite{Olsen2023}. The training data set is $\mathcal{X} = \{\boldsymbol{x}^{[i]}, y^{[i]}\}_{i=1}^{N_{\text{train}}}$ with $N_\text{train} = 1000$. Here $\x^{[i]} \sim\mathcal{N}_{8}(\0, \Sigma)$, where $\Sigma_{jl} = \rho^{\abs{j-l}}$ for $\rho = 0.5$. While the response $y^{[i]} = \beta_0 + \sum_{j=1}^{M}\beta_j\cos(x_j^{[i]}) + \gamma_1g(x_1^{[i]}, x_2^{[i]}) + \gamma_2g(x_3^{[i]}, x_4^{[i]}) + \varepsilon^{[i]}$, where $g(x_j,x_k) = x_jx_k + x_jx_k^2 + x_kx_j^2$, $\boldsymbol{\beta} = \{1.0,  0.2, -0.8, 1.0, 0.5, -0.8, 0.6, -0.7, -0.6\}$, $\boldsymbol{\gamma} = \{0.8, -1.0\}$, and $\varepsilon^{[i]} \sim \mathcal{N}(0, 1)$. The corresponding predictive model $f(\x)$ is a GAM with splines for the nonlinear terms and tensor product smooths for the nonlinear interaction terms. We create $N_\text{test} = 250$ test observations by the same data-generating procedure and explain the corresponding predictions made by $f$.

A common evaluation criterion in XAI is the mean absolute error (MAE) between the true and estimated Shapley values averaged over all test observations and features, see \cite{aas2021explaining, Olsen2023, Olsen2022}. That is,  $\operatorname{MAE} = \operatorname{MAE}_{\phi}(\text{method } \texttt{q}) =
\tfrac{1}{N_\text{test}} \sum_{i=1}^{N_\text{test}} \tfrac{1}{M} \sum_{j=1}^M |\phi_{j, \texttt{true}}(\boldsymbol{x}^{[i]}) - \hat{\phi}_{j, \texttt{q}}(\boldsymbol{x}^{[i]})|$. The true Shapley values are generally unknown, but we can compute them with arbitrary precision in our setup as we know the true data-generating process. The MAE is suitable to measure the average precision of a method, but it tells us nothing about the spread in the errors for the different test observations. Uncovering systematized patterns in the errors is of high interest in the Shapley value explanation setting as the explanations are used on an individual basis.

\begin{figure}[!t]
	\centering
	\centerline{\includegraphics[width=1\textwidth]{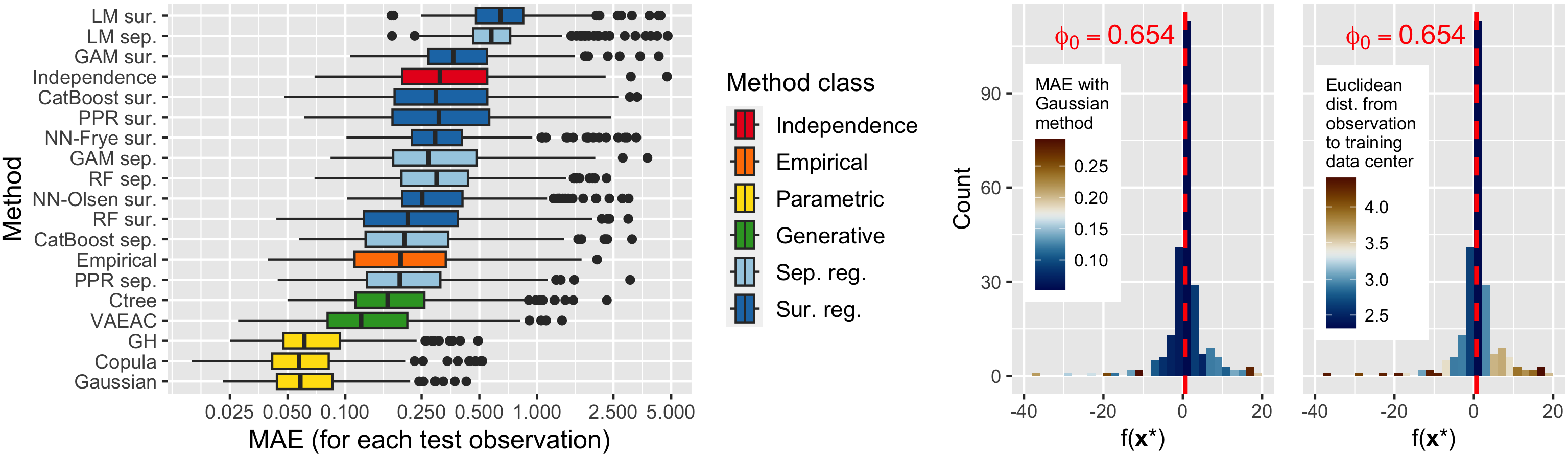}}
	\captionsetup{belowskip=-3pt} 
		\captionsetup{aboveskip=5pt} 
	\caption{{\small Left: Boxplot of the MAE between the true and estimated Shapley values in the \texttt{gam\textunderscore more\textunderscore interactions} experiment, ordered based on overall MAE. Right: Histograms of the explained predictions $f(\x^*)$ with color indicating the corresponding MAE using the \Gaussian\ method or Euclidean distance.}}
	\label{fig:Simulation:MAE_boxplot_histogram}
\end{figure}

\begin{figure}[!t]
	\centering
	\centerline{\includegraphics[width=1\textwidth]{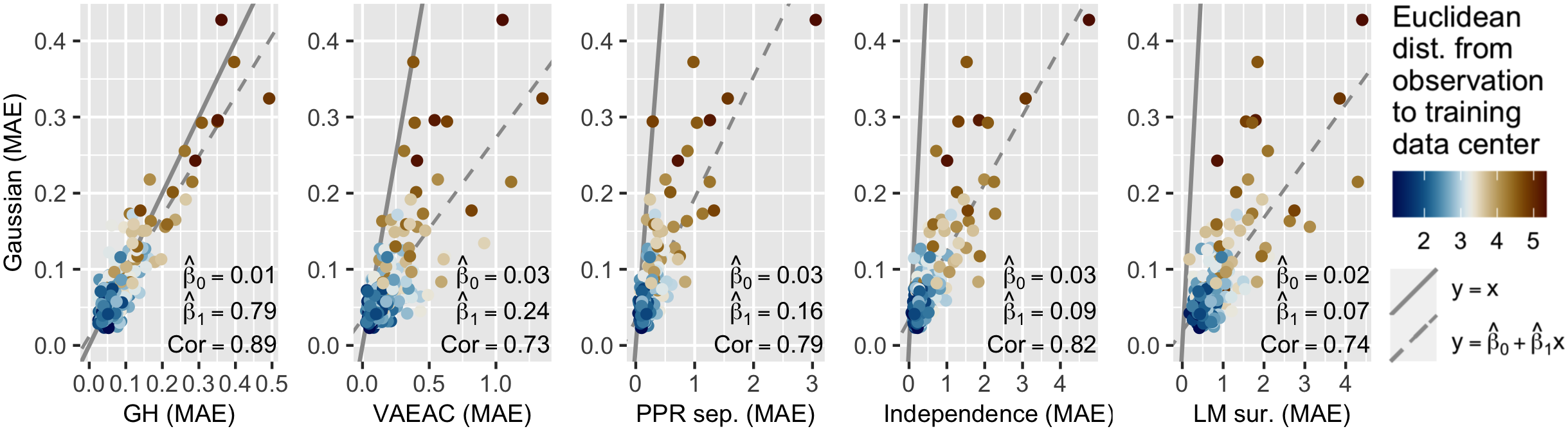}}
	\captionsetup{belowskip=-17pt} 
	\captionsetup{aboveskip=5pt} 
	\caption{{\small Plots of the MAE for each test observation for different pairs of methods.}}
	\label{fig:Simulation:MAE_Individuals}
\end{figure}

In \Cref{fig:Simulation:MAE_boxplot_histogram}, we see that the \parametric\ methods, which are all able to model the Gaussian distribution, obtain the lowest MAE and, therefore, the most precise Shapley value explanations. However, we see several outliers for most methods, that is, greater than the upper quartile plus $1.5$ times the interquartile range. In \Cref{fig:Simulation:MAE_Individuals}, we see that it is the same test observations that yield large errors for all methods. Furthermore, we see a clear pattern in the errors when we color encode the test observations based on their Euclidean distance to the empirical center of the training data distribution. Note that using the empirical mean as the center does not apply to multimodal data. 

The predictions with the largest Shapley value explanation errors correspond to the test observations furthest away from the center of the training data. This is not surprising from a statistical point of view as the estimation methods have little to no training data to learn the feature dependence structures in these outer regions. Making a correct parametric assumption about the data reduces the errors compared to the more flexible methods, but the errors are still larger compared to test observations closer to the training data center.

\begin{figure}[!t]
	\centering
	\centerline{\includegraphics[width=1\textwidth]{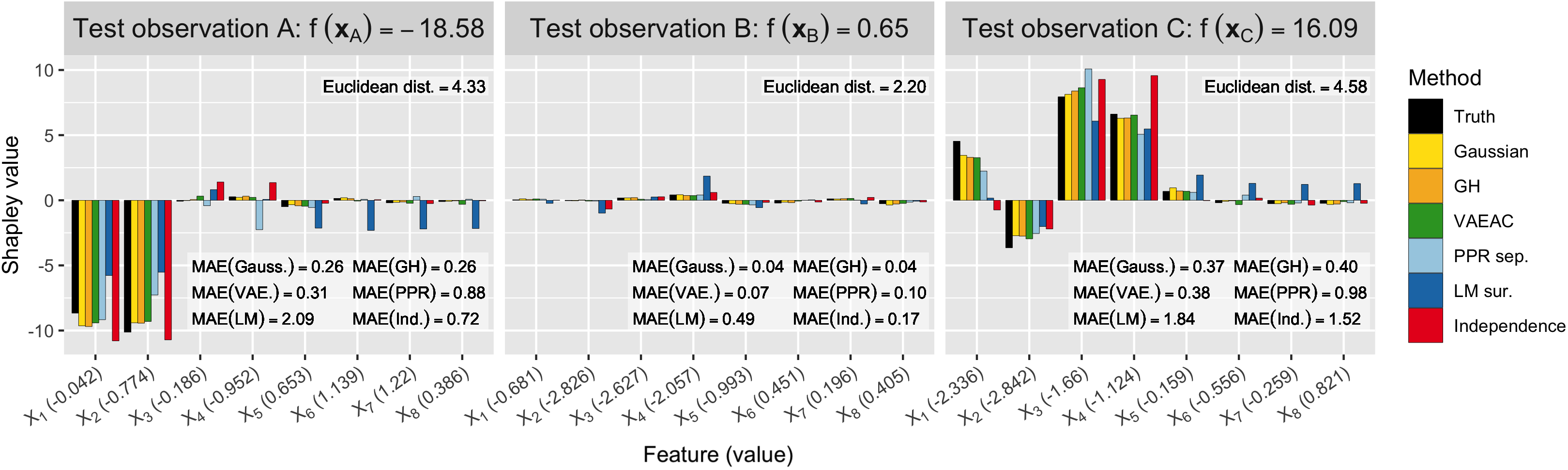}}
	\captionsetup{belowskip=-15pt} 
	\captionsetup{aboveskip=3pt} 
	\caption{{\small Plots of the true and estimated Shapley values for three test observations.}}
	\label{fig:Simulation:Shapley_values}
\end{figure}

The left histogram in \Cref{fig:Simulation:MAE_boxplot_histogram} illustrates that observations $\x^*$ with a large $|f(\x^*) - \phi_0|$ yield larger mean absolute errors. This tendency is natural as the magnitude of the Shapley values $\bphi^*$ has to increase since $f(\x^*) = \phi_0 + \sum_{j=1}^M\phi_j^*$. Meaning that the scale of the Shapley values changes based on $\x^*$, as we illustrate in \Cref{fig:Simulation:Shapley_values} for three test observations with predicted responses below, close, and above $\phi_0$. Thus, the relative Shapley value errors can be larger for observations with predictions close to $\phi_0$, but this has a minimal impact on the overall MAE due to its absolute and not relative scale. Scaling the MAE by $|f(\x^{[i]}) - \phi_0|^{-1}$, for $i = 1,2,\dots,N_\text{test}$, leads to problems as the MAE will blow up to infinity for $f(\x^{[i]}) \approx \phi_0$ and is not defined for $f(\x^{[i]}) = \phi_0$.

In \Cref{fig:Simulation:Shapley_values}, we see that most methods find the true influential features. Thus, if the Shapley values are only used to roughly uncover the important features and rank them, then the practitioner can be less careful than if the estimated Shapley values are directly used in further analysis. For the latter, the practitioner should consider the location of $\x^*$.

\vspace{-1.4ex} \section{Conclusion} \vspace{-1.4ex}
We have demonstrated and discussed that Shapley value explanations are less precise for test observations in the outer regions of the training data, w.r.t.\ the MAE evaluation criterion, and argued that practitioners should be more careful when applying these observations' Shapley value explanations.

\vspace{-0.5ex}
\acknowledgements{I want to thank my supervisors, Ingrid Glad, Martin Jullum, and Kjersti Aas, for their guidance. This work was supported by The Norwegian Research Council 237718 through the research center BigInsight.}
\vspace{-2ex}
\scriptsize{
}
\end{document}